\documentclass[10pt,twocolumn,letterpaper]{article}

\usepackage{wacv}
\usepackage{times}
\usepackage{epsfig}
\usepackage{graphicx}
\usepackage{amsmath}
\usepackage{amssymb}

\def\wacvPaperID{808} 

\wacvfinalcopy 

\ifwacvfinal
\usepackage[breaklinks=true,bookmarks=false]{hyperref}
\else
\usepackage[pagebackref=true,breaklinks=true,colorlinks,bookmarks=false]{hyperref}
\fi

\ifwacvfinal
\else
\fi

\begin{document}

\title{SMPLpix: \\ Neural Avatars from 3D Human Models} 



\author{Sergey Prokudin \thanks{work was done during internship at Amazon} \\
Max Planck Institute for Intelligent Systems\\
T\"ubingen Germany\\
{\tt\small sergey.prokudin@tuebingen.mpg.de}
 
\and
Michael J. Black\\
Amazon\\
T\"ubingen Germany\\
{\tt\small mjblack@amazon.com}



\and
Javier Romero\\
Amazon\\
Barcelona Spain\\\
{\tt\small javier@amazon.com}
}

\maketitle
\thispagestyle{empty}

\begin{abstract}
Recent advances in deep generative models have led to an unprecedented level of realism for synthetically generated images of humans. However, one of the remaining fundamental limitations of these models is the ability to flexibly control the generative process, e.g.~change the camera and human pose while retaining the subject identity. At the same time, deformable human body models like SMPL \cite{loper2015smpl} and its successors provide full control over pose and shape, but rely on classic computer graphics pipelines for rendering. Such rendering pipelines require explicit mesh rasterization  that (a) does not have the potential to fix  artifacts or lack of realism in the original 3D geometry and (b) until recently, were not fully incorporated into deep learning frameworks. In this work, we propose to bridge the gap between classic geometry-based rendering and the latest generative networks operating in pixel space. We train a network that directly converts a sparse set of 3D mesh vertices into photorealistic images, alleviating the need for traditional rasterization mechanism. We train our model on a large corpus of human 3D models and corresponding real photos, and show the advantage over conventional differentiable renderers both in terms of the level of photorealism and rendering efficiency.

\end{abstract}

\section{Introduction}

\begin{figure*}[t]
\begin{center}
  \includegraphics[clip,width=\textwidth]{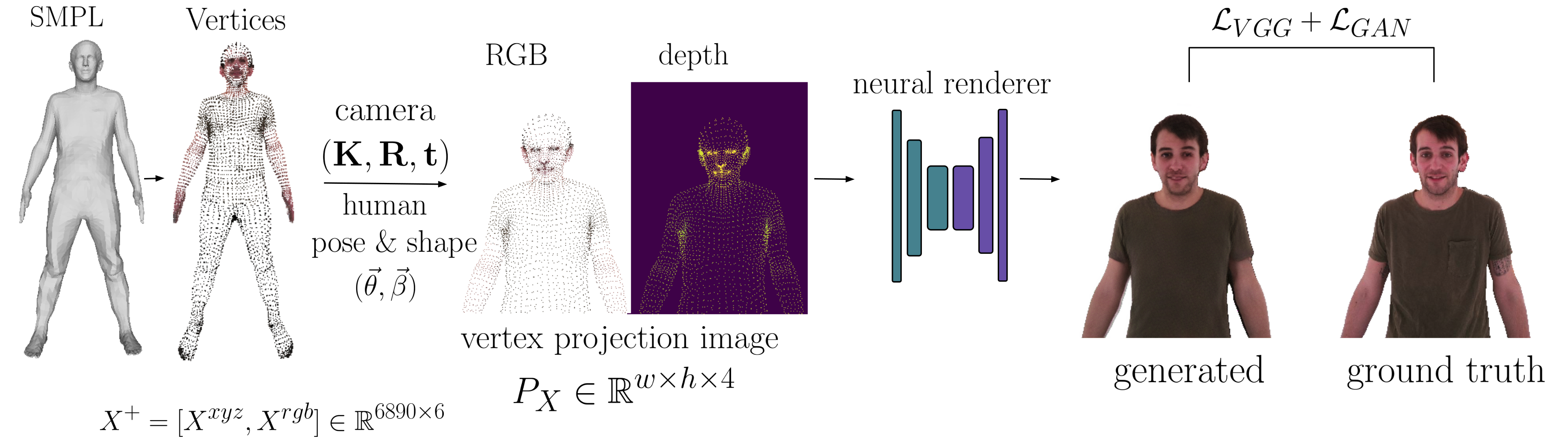}
\end{center}
 \caption{\textit{SMPLpix Neural Rendering Pipeline.} Training SMPLpix requires a set of 3D vertices with the corresponding RGB colors as input $X^+$, along with ground truth camera parameters $(\mathbf{K}, \mathbf{R}, \mathbf{t})$. Our training data is obtained by registering a SMPL model to 3D scans. Using SMPL also allows us to control the coordinates of $X^+$ via a small set of pose parameters $\theta$.
 RGB-d training images are created by projecting the vertices, $X^+$ onto an image plane using a camera model. This image is then fed into a UNet-type network that reconstructs surfaces from projected vertices \emph{directly in the pixel space}. It is trained to minimize a combination of perceptual and adversarial losses w.r.t.~the ground truth image. Once trained, this neural rendering module generalizes to unseen subjects $X^+$, body poses $\theta$ and camera parameters $(\mathbf{K}, \mathbf{R}, \mathbf{t})$.
 }
\label{fig:fig1}
\vspace{-5mm}
\end{figure*}

Traditional graphics pipelines for human body and face synthesis benefit from explicit, parameterized, editable representations of 3D shape and the ability to control pose, lighting, material properties, and the camera, to animate 3D models in 3D scenes.
While photorealism is possible with classical methods, this typically comes at the expense of complex systems to capture detailed shape and reflectance or heavy animator input.
In contrast, recent developments in deep learning and the evolution of graphics processing units are rapidly bringing new tools for human modeling, animation and synthesis.  
Models based on generative adversarial networks \cite{goodfellow2014generative} reach new levels of realism in synthesizing human faces \cite{karras2019style,karras2019analyzing} and various models can repose humans \cite{chan2019everybody}, swap identities and appearance, etc. 

While promising, particularly in terms of their realism, these new ``neural" approaches to synthesizing humans have several drawbacks relative to classical methods.
Specifically, a key advantage of classical graphics methods \cite{pharr2016physically} is the ability to fully and flexibly control the generative process, e.g. change the camera view, the light or even the pose or shape of the subject. 
These methods, however, have two main limitations relative to learning-based image synthesis. First, until recently \cite{kato2018neural,liu2019soft}, rendering engines were not fully integrated into deep learning pipelines. Second, explicit mesh-based rendering methods are limited when it comes to rendering complex, high-frequency geometry (e.g. hair or fur, wrinkles on clothing, etc.) and dealing with complex, changing, topology.
The future of graphics is likely a synthesis of classical and neural models, combining the best properties of both.
Here we make a  step in this direction by combining the parameterized control of 3D body shape and pose with neural point-based rendering, which replaces the classical rendering pipeline.

Point-based rendering has a long history in computer graphics \cite{gross2011point,kobbelt2004survey}. 
Recently, point-based rendering has been successfully coupled with the neural network pipeline via learning per-point neural descriptors that are interpreted by the neural renderer \cite{aliev2019neural}. This approach produces photo-realistic novel views of a scene from a captured point cloud. However, this pipeline has been demonstrated for rendering static scenes with dense point clouds as inputs, with the need of re-learning point descriptors for every novel scene.

Our approach is influenced by \cite{aliev2019neural} and \cite{martin2018lookingood}. However, along with the technical novelties and simplifications we describe in the follow-up sections, our main aim is to extend these approaches to enable efficient rendering of human avatars \emph{under novel subject identities and human poses}. We accomplish this by introducing SMPL \cite{loper2015smpl}, a deformable 3D body model, into the neural rendering loop. This provides us full control over body pose and shape variation. However, instead of relying on mesh connectivity for explicit rendering, we simply use mesh vertices and their colors projected onto the image plane as inputs for the neural rendering module. 
This provides the benefits of a parameterized body model, while greatly improving the rendering quality, without the complexity of classical methods.

The overall pipeline, called \textit{SMPLpix}, is outlined in Figure \ref{fig:fig1}. During training, our framework operates on the data obtained from a commercially available 3D scanner \cite{treedys}. The SMPL model is registered to the raw scans \cite{bogo2014faust,loper2015smpl}; other parametric models can be used in principle \cite{joo2018total,pavlakos2019expressive}. The result of this process is a set of mesh vertices $X \in \mathbb{R}^{6890\times3}$, the RGB color of each vertex, and the body pose parameters $\theta$.
It is important to mention that the registration process has inherent limitations like fitting hair (due to the irregularity of hair and low resolution of the SMPL model) or fitting clothing (due to the form-fitting topology of SMPL). The advantage of using the registered vertices over raw scans, however, is that we can control the pose of the vertices $X$ by varying a small set of inferred pose parameters $\theta$. We project the vertices of the body model using ground truth scanner camera locations $(\mathbf{K}, \mathbf{R}, \mathbf{t})$ and obtain an RGB-d image of the projected vertices. This image is processed by a UNet-like neural rendering network to produce the rasterized output RGB image that should match the ground truth image from a scanner camera. At test time, we are given novel mesh vertices $X$, their colors, body poses $\theta$ and camera locations $(\mathbf{K}, \mathbf{R}, \mathbf{t})$.
Note that this input can also come from the real images using methods like \cite{alldieck2019learning}.

\textbf{Intuition.} Our proposed method can be seen as a middle ground between mesh-based and point-based renderers. While we use the structured nature of mesh vertices to control the generative process, 
we ignore the mesh connectivity  and treat vertices simply as unstructured point clouds. Compared with explicit mesh rasterization, the main advantage of this vertices-as-points approach, along with its computational and conceptual simplicity, is the ability of the trained neural renderer to reproduce complex high frequency surfaces \emph{directly in the pixel space}, as we will show in the experimental section. Our approach is also potentially applicable in cases when no explicit mesh connectivity information is available whatsoever and only a set of 3D anchor points is given.

\textbf{Contributions.} The proposed work offers the following contributions:
\begin{itemize}
    \item \emph{Deep controlled human image synthesis}: apart from the classic mesh-based renderers, to the best of our knowledge, the presented approach is the first one that can render novel human subjects under novel poses and camera views. The proposed framework  produces photo-realistic images with complex geometry that are hard to reproduce with these classic renderers;
    \item \emph{Sparse point set neural rendering}: we show how popular image-to-image translation frameworks can be adapted to the task of translating a sparse set of 3D points to RGB images, combining several steps (geometric occlusion reasoning, rasterization and image enhancement) into a single neural network module.
\end{itemize}

\section{Related work}
\label{sec:related}

Our method is connected to several broad branches of 3D modeling and image synthesis techniques. Here we focus on the most representative work in the field.

\textbf{3D human models.} Our method is based on the idea of modeling humans bodies and their parts via deformable 3D models \cite{anguelov2005scape,blanz1999morphable,joo2018total}, and in particular SMPL \cite{loper2015smpl}. Such models are controllable (essential for graphics) and interpretable  (important for analysis). Extensions of SMPL exist that also model hands \cite{romero2017embodied}, faces \cite{li2017learning,pavlakos2019expressive} and  clothing \cite{ma2019learning}. Separate models exist for capturing and modeling clothing, wrinkles, and hair \cite{hu2017simulation,zhou2018hairnet}. 
While powerful, 
rendering such models requires high-quality textures and accurate 3D geometry, which can be hard to acquire.
Even then, the resulting rendered images may look smooth and fail to model details that are not properly captured by the model or surface reconstruction algorithms.

\textbf{Neural avatars.} Recently, a new work focuses learning to render high-fidelity digital avatars \cite{lombardi2018deep,shysheya2019textured,thies2019deferred,wei2019vr}. While these works provide a great level of photo-realism, they are mostly tailored to accurately \emph{modeling a single subject}, and part or the whole system needs to be retrained in case of a new input. In contrast, our system is trained in a multi-person scenario and can render unseen subjects at test time. Another advantage is that it takes a relatively compact generic input (a set of 3D mesh vertices and their RGB colors) that can be also inferred from multiple sources at test time, including from real-world images \cite{alldieck2019learning}.

\textbf{Pixel-space image translation and character animation.} The second part of our system, neural human renderer, is based on the recent success of pixel-to-pixel image translation techniques \cite{esser2018variational,isola2017image,wang2018high}. Two particular variations of this framework have the most resemblance to our model. 
First, \cite{chan2019everybody} uses a set of sparse body keypoints (inferred from a source actor) as input to produce an animated image sequence of a target actor. However, as with the neural avatars discussed above, the system needs to be retrained in order to operate on a novel target subject. Our work also resembles the sketch-to-image translation regime, where an edge image is used in order to produce a photo-realistic image of the person's head \cite{zakharov2019few} or generic objects
\cite{chen2018sketchygan}. Our approach can also be viewed as translating a sparse set of key points into an image. However, our keypoints  come from a structured 3D template and therefore convey more information about the rendered subject appearance; since they exist in 3D, they can be projected to an image plane under different camera views. Finally, another advantage of using SMPL  topology as input to our image translation framework is its non-uniform vertex density according to region importance (i.e.~faces and hands are more densely sampled). This makes detailed rendering of these regions easier, without the need for a specific attention mechanism in the neural renderer itself.

\textbf{Differentiable mesh (re-)rendering.} There are several available solutions that incorporate the mesh rendering step into fully differentiable learning pipelines \cite{kato2018neural,liu2019soft,loper2015smpl}. However, these methods follow a different line of work: they aim at constructing better gradients for the mesh rasterization step, while keeping the whole procedure of mesh face rendering and occlusion reasoning deterministic. This applies also to a soft rasterizer \cite{liu2019soft} that substitutes the discrete rasterization step with a probabilistic alternative. While this proves useful for the flow of gradients, the rendering procedure still lacks the flexibility that would allow it to fix artifacts of the original input geometry. One potential solution is to enhance the produced incomplete noisy renders by the additional neural re-rendering module \cite{liu2020neural,martin2018lookingood}. Our framework can be seen as the one that combines standard mesh rendering step with a follow-up neural image enhancement into one task-specific neural rendering module. Considering the original target application of \cite{martin2018lookingood}, another potential advantage of our framework for online conferencing is the reduced amount of data that needs to be transferred over the network channel to produce the final image. 

\textbf{Point-based rendering.} Point-based rendering \cite{gross2011point,kobbelt2004survey,levoy1985use,pfister2000surfels,rusinkiewicz2000qsplat} offers a well-established, scalable alternative to rendering scenes that can be hard to model with surface meshing approaches. We take inspiration from these methods, however, we substitute the fixed logic of rendering (e.g.~surfel-based \cite{pfister2000surfels}) with a neural module in order to adapt to sparse point sets with highly non-uniform densities, as well as to generate photorealistic pixel-space textures. 

\textbf{Rendering from deep 3D descriptors.} Another promising direction for geometry-aware image synthesis aims to learn some form of deep 3D descriptors from a 2D or 3D inputs \cite{aliev2019neural,lombardi2019neural,sitzmann2019deepvoxels,sitzmann2019srns}. These descriptors are processed by a trainable neural renderer to generate novel views. These methods, however, are limited when it comes to controlling the generative process; shapes are represented as voxels \cite{lombardi2019neural,sitzmann2019deepvoxels}, unstructured point clouds \cite{aliev2019neural} or neural network weights \cite{sitzmann2019srns}.
This makes parameterized control of human pose difficult. 

\textbf{Neural point-based graphics.} The closest work to ours is \cite{aliev2019neural}. An obvious difference with respect to this work is that our input comes from a deformable model, which allows us to modify the render in a generative and intuitive way.
Moreover, our model contains two additional differences. First, our inputs are considerably sparser and less uniform than the point clouds considered in \cite{aliev2019neural}. Second, instead of point neural descriptors that need to be relearned for every novel scene or subject, our rendering network obtains the specific details of a subject through the RGB colors it consumes as input \emph{at test time}. This alleviates the need for retraining the system for every novel scene. 

In summary, SMPLpix fills an important gap in the literature, combining the benefits of parameterized models like SMPL with the power of neural rendering. The former gives controllability, while the latter provides realism that is difficult to obtain with classical graphics pipelines. 


\section{Method}

\label{sec:method}

As is common in deep learning systems, our system has two key parts: the data used for training our model, and the model itself. We describe those two parts in the following sections.

\subsection{Data}
\label{subsec:smpl}

\textbf{Scans.} Our renderer transforms sparse RGB-D images obtained from the 2D projections of SMPL~\cite{loper2015smpl} vertices.
We take a supervised training approach with ground-truth images that correspond to the projected vertices of the SMPL model.
Although it would be ideal to collect such a dataset from images in the wild, the inaccuracies in methods that infer SMPL bodies from images (e.g.~\cite{kanazawa2018end}) currently make this data ineffective. Instead, we use scan data collected in the lab. To that end, we collected more than a thousand scans with a commercially available 3D scanner (Treedy's, Brussels, Belgium \cite{treedys}) and photogrammetry software (Agisoft Photoscan~\cite{agisoft}). 
This results in raw 3D point clouds (\emph{scans}) $S \in \mathbb{R}^{M \times 6}, M \approx 10^6$, representing the body geometry, together with camera calibration $(\mathbf{K}, \mathbf{R}, \mathbf{t})$ compatible with a pinhole camera model. Note that the subjects are scanned in a neutral A-pose. Unlike most other image generation methods, this is not a problem for our system since the strong guidance provided by the input images prevents our method from overfitting to the input pose, as it can be seen in Section~\ref{subsec:qualitative}.

\textbf{Registrations.}
While these scans could potentially undergo a rendering process like~\cite{aliev2019neural}, it would not be possible to deform them in a generative manner, i.e. changing their shape or pose. To achieve that, we transform those unstructured point clouds into a set of points $X \in  \mathbb{R}^{N \times 3}, N=6890$ with fixed topology that correspond to a reshapeable and reposeable model, SMPL \cite{loper2015smpl}. 
In its essence, SMPL is a linear blend skinned (LBS) model that represents the observed body vertices $X$ as a function of identity-dependent and pose-dependent mesh deformations, driven by two corresponding compact sets of shape $\Vec{\beta} \in \mathbb{R}^{10}$ and pose $\Vec{\theta} \in \mathbb{R}^{72}$ parameters:
%
\begin{eqnarray}
X = W(T_{P}(\Vec{\beta}, \Vec{\theta}), J(\Vec{\beta}), \Vec{\theta}, \mathcal{W}), \label{eq:smpl_lbs}\\
T_{P}(\Vec{\beta}, \Vec{\theta}) = \mathbf{\bar{T}} + B_S(\Vec{\beta}) + B_P(\Vec{\theta}),
\label{eq:smpl_temp}
\end{eqnarray}
where $T_{P}(\Vec{\beta}, \Vec{\theta})$  models shape and pose dependent deformation of the template mesh in the canonical T pose via linear functions $B_S$ and $B_P$, and $W$ corresponds to the LBS function that takes the T-pose template $T_{P}$, set of shape-dependent $K$ body joint locations $J(\Vec{\beta}) \in \mathbb{R}^{3K}, K=23$ and applies the LBS function $W$ with weights $\mathcal{W}$ to produce the final posed mesh. We refer to the original publication \cite{loper2015smpl} for more details on the SMPL skinning function.
Note that other versions of deformable 3D models \cite{anguelov2005scape,joo2018total} or topologies could be used, including the ones that additionally model hands and faces \cite{li2017learning,pavlakos2019expressive,romero2017embodied}, as well as clothing deformations \cite{ma2019learning}. 
In fact, in Section~\ref{subsec:results} we show experiments with two topologies of different cardinality.

The SMPL registration process optimizes the location of the registration vertices and the underlying model, so that the distance between the point cloud and the surface entailed by the registration is minimized, while the registration vertices remain close to the optimized model. It is inspired by the registration in~\cite{bogo2014faust} although the texture matching term is not used.
It is worth emphasizing that these registrations, as in~\cite{bogo2014faust}, can contain details about the clothing of the person since their vertices are optimized as free variables. This does not prevent us from reposing those subjects after converting them into SMPL templates $\mathbf{\bar{T}}^*$ through unposing, as explained and shown in Section~\ref{subsec:qualitative}. However, these extra geometric details are far from perfect, e.g.~they are visibly wrong in the case of garments with non-anthropomorphic topology, like skirts.

\textbf{Color.}
Finally, the registered mesh is used in Agisoft Photoscan together with the original image and camera calibration to extract a high-resolution texture image $I_{tex} \in \mathbb{R}^{8192 \times 8192 \times 3}$. This texture image is a flattened version of the SMPL mesh, in which every 3D triangle in SMPL corresponds to a 2D triangle in the texture image. Therefore, each triangle contains thousands of color pixels representing the appearance of that body portion. These textures can be used directly by the classic renderer to produce detailed images, as can be seen in Section~\ref{subsec:results}. Although it would be possible to exploit the detail in those textures by a neural renderer, that would slow it down and make it unnecessarily complex. Instead, we propose to use the sparse set of colors $X^c \in [0,1]^{6890\times 3}$ sampled at the SMPL vertex locations. These colors can be easily extracted from the texture image, since they are in full correspondence with the mesh topology.  

\textbf{Projections.}
Having an input colored vertex set $X^+=[X, X^c] \in \mathbf{R}^{6890 \times 6}$ and camera calibration parameters $(\mathbf{K}, \mathbf{R}, \mathbf{t})$, we obtain image plane coordinates for every vertex $x \in X$ using a standard pinhole camera model \cite{hartley2003multiple}:
\begin{eqnarray}
\begin{pmatrix}u\\v\\d\end{pmatrix} = \mathbf{K}(\mathbf{R}x + \mathbf{t}).
\label{eq:camera}
\end{eqnarray}

Next, we form an RGB-D vertex projection image.
The projection image $P_X \in \mathbb{R}^{w \times h \times 4}$ is initialized to a value that can be identified as background by its depth value. Since depth values collected in the scanner have a range between $0.1$ and $0.7$ meters, a default value of $1$ is used to initialize both RGB and depth in $P_X$.
Then, for every vertex $x \in X$, its image plane coordinates $(u, v, d)$ and color values $(r, g, b) \in X^c$ we assign:
\begin{eqnarray}
P_X[\lfloor u \rfloor, \lfloor v \rfloor] = (r, g, b, d).
\label{eq:projection}
\end{eqnarray}

In order to resolve collisions during the projection phase (\ref{eq:projection}), when different vertices from $X$ end up sharing the same pixel-space coordinates $\lfloor u \rfloor, \lfloor v \rfloor$, we sort the vertices according to their depth and eliminate all the duplicate consecutive elements of the depth-wise sorted array of $\lfloor u \rfloor, \lfloor v \rfloor$ coordinates of X. Note that, since the number of vertices is much smaller than the full resolution of the image plane, these collisions rarely happen in practice. 

The whole vertex projection operation (\ref{eq:camera})-(\ref{eq:projection}) can be easily and  efficiently implemented within modern deep learning frameworks \cite{paszke2017automatic} and, therefore, seamlessly integrated into bigger pipelines.


\subsection{Neural rendering}
\label{subsec:rasterization}

Given our training data consisting of pairs of RGB-D projection images $P_X$ and segmented output images $I_X$, we train a UNet-type \cite{ronneberger2015u} neural network $G$ with parameters $\Theta$ to map initial point projections to final output images:
\begin{eqnarray}
G_{\Theta}: P_X \rightarrow I_X  .
\label{eq:unet}
\end{eqnarray}

In our experiments, we use one of the  publicly available UNet architecture designs~\cite{unetpytorch}, to which we apply only minor changes to adapt it to our types of input and output. The network consists of 4 layers of $downconv$ and  $upconv$ double convolutional layers [Conv2d, BatchNorm, ReLU] $\times 2$, with convolutional layers having the kernel size of 3. In case of $downconv$, this double convolutional layer is preceded by max pooling operation with kernel size 2; in case of $upconv$, it is preceded by bilinear upsampling and concatenation with the output of a corresponding $downconv$ layer. In general, the particular design of this module can be further optimized and tailored to a specific target image resolution and hardware requirements; we leave this optimization and further design search for a future work. 

Having the ground truth image $I_{gt}$ for a given subject and camera pose, we optimize our rendering network $G_{\Theta}$ for the weighted combination of perceptual VGG-loss \cite{johnson2016perceptual}, multi-scale, patch-based GAN loss and  feature matching GAN loss \cite{wang2018high} in two stages.

During the first stage (100 epochs), we train the model with Adam (learning rate set to 1.0e-4) and batch size 10 by minimizing the $L1$ loss between VGG activations:
\begin{multline}
L_{VGG}(I_{gt}, I_{X}) =\\ \sum^5_{i=0}{\frac{1}{2^{(5-i)}} ||f_{VGG}^{(i)}(I_{gt}) - f_{VGG}^{(i)}(I_{X})||_1},
\label{eq:vgg_loss}
\end{multline}
where $f_{VGG}^{(i)}(I)$ are activations at layer $i$ and  $f_{VGG}^{(0)}(I) = I$.

During the second stage (100 epochs), we restart Adam with learning rate 1.0e-5 and include a combination of multi-scale GAN and feature-matching losses identical to the ones in \cite{wang2018high}:
\begin{multline}
L(I_{gt}, I_{X}) = L_{VGG}(I_{gt}, I_{X}) \\+ \min\limits_{G}\Big[\max\limits_{D_1, D_2, D_3}\sum_{k=1,2,3}{L_{GAN}(G, D_k)}  \\+ 0.1 * \sum_{k=1,2,3}{L_{FM}(G, D_k)} \Big] .
\end{multline}

Implicitly, the network $G_{\Theta}$ is learning to accomplish several tasks. First, it needs to learn some form of geometric reasoning, i.e. to ignore certain projected vertices based on their depth values. In that sense, it substitutes fixed-logic differentiable mesh rendering procedures \cite{kato2018neural} with a flexible, task-specific neural equivalent. Second, it needs to learn how to synthesize realistic textures based on sparse supervision provided by the projected vertices, as well as to hallucinate  whole areas not properly captured by the 3D geometry, e.g. hair and clothing, to match the real ground truth images.  Therefore, we believe that this approach could serve as a potentially superior (in terms of acquired image realism), as well as easier to integrate and computationally flexible, alternative to the explicit fixed differentiable mesh rasterization step of \cite{kato2018neural}.

\begin{figure*}[t!]
\begin{center}
  \includegraphics[clip,width=\textwidth]{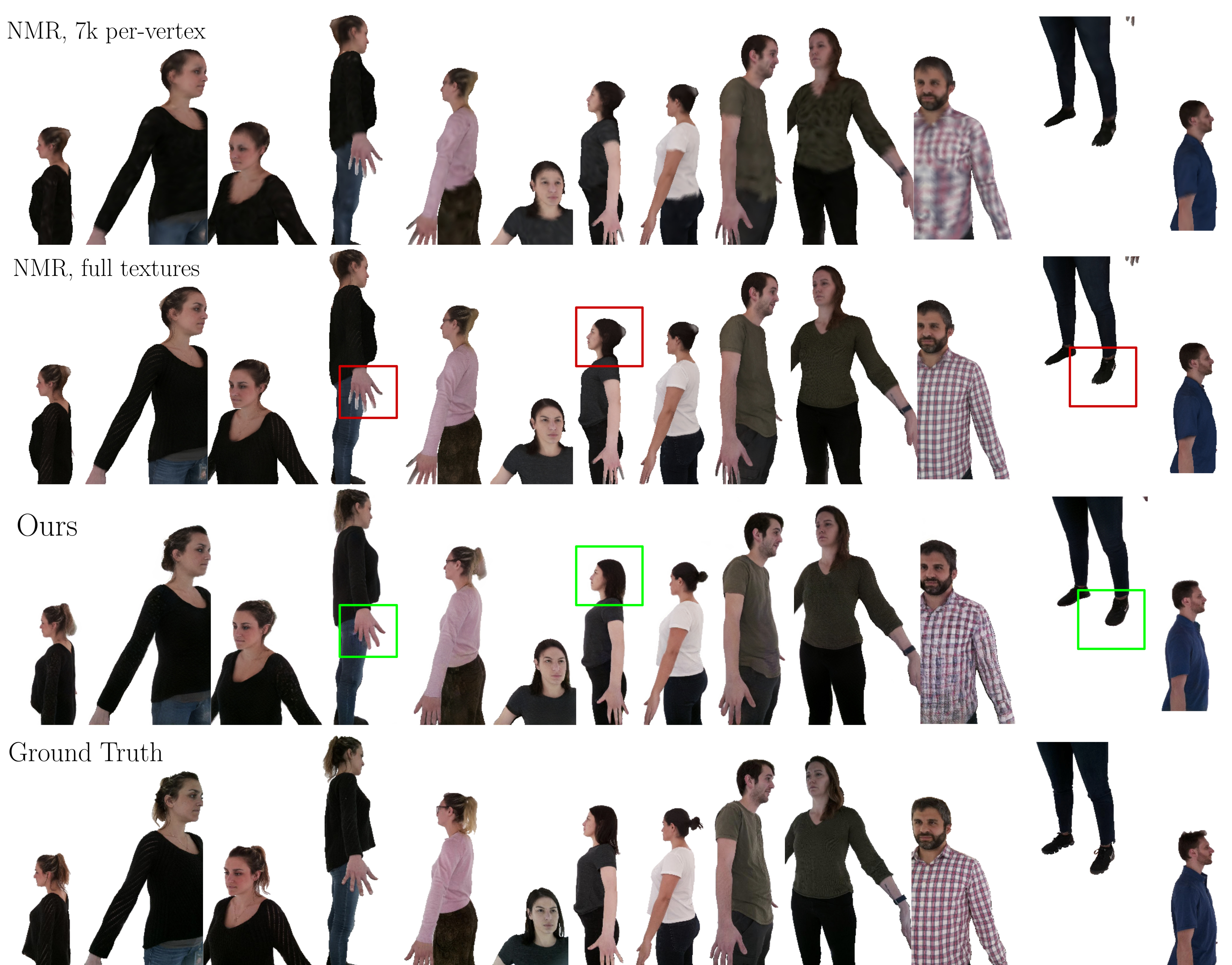}
\end{center}
 \caption{\textit{Qualitative comparison between neural mesh renderer \cite{kato2018neural} and SMPLpix (27k vertices) on novel subjects and camera poses (zoom in for details).} Compared to a standard mesh renderer, our model can fix texture and geometry  artefacts (toe and finger regions) and generate high frequency details (hair and cloth wrinkles), while remaining conceptually simple (point projections as the main 3D geometry operator) and efficient in terms of utilized data and inference time.}
\label{fig:fig2}
 \vspace{-5mm}
\end{figure*}

\section{Experiments}
\subsection{Data details}
\label{subsec:data}

Accurately captured, well-calibrated data is essential for the proposed approach in its current form. We use 3D scans of 1668 subjects in casual clothing. The subjects are diverse in gender, body shape, age, ethnicity, as well as clothing patterns and style. For each subject, we select 20 random photos from among the 137 camera positions available in the scanner camera rig. We use 1600 subjects for training and 68 subjects for test, which forms training and test sets of $32000$ and $1360$ images correspondingly.
We use the image resolution of size $410 \times 308$ during all the experiments. Of 68 test subjects, 16 gave their explicit consent for their images to be used in the present submission. We use these test subjects for the qualitative comparison in the paper, while the full test set is used for the quantitative evaluation.

\subsection{Quantitative experiments}
\label{subsec:results}

We compare our system with other renderers that can generate images of reshapeable and reposeable bodies. This limits the other methods to be classic rendering pipelines, since, to the best of our knowledge, no other deep learning model offers this generative behaviour. It is important that the renderers support automatic differentiation, since our ultimate goal includes integrating the renderer with a fully differentiable learning system. With these two constraints, we compare with the neural mesh renderer introduced in~\cite{kato2018neural}, in its popular PyTorch re-implementation~\cite{pytorchnr}.

\textbf{Metrics.} We compare SMPLpix against different versions of classic renders implemented with~\cite{pytorchnr} according to two different quantitative metrics popular in image generation and super-resolution: peak signal-to-noise ratio (PSNR, \cite{hore2010image}) and learned perceptual image patch similarity (LPIPS,~\cite{zhang2018unreasonable}). PSNR is a classic method, while LPIPS has gained popularity in recent works for being more correlated with the perceptual differences. We should note that the field of quantitative perceptual evaluation is still an area of research, and no metric is perfect. Therefore, we also provide qualitative results in the next section.

\textbf{Baseline variants.} For \cite{kato2018neural}, we use the following rendering variants. First, we render the mesh with exactly the same information available to our SMPLpix rendering pipeline, i.e.~only 1 RGB color per vertex\footnote{Technically, since~\cite{pytorchnr} does not support per-vertex color rendering, this has to be achieved by performing linear interpolation between the vertex colors in their per-triangle texture space}. Next, we use the much more information-dense option of texture images $I_{tex}$. To optimise the inference time of \cite{kato2018neural}, we do not utilise the full extensive 8k textures, but rather search for the optimal downscaled version of the texture image, at which no further improvement in terms of PSNR and LPIPS were observed (Table \ref{table:results}, row 2). 
Since our method can be topology agnostic, we perform these comparisons for two topologies: the native SMPL topology of 6890 vertices (noted as $7k$) and an upsampled version with a higher vert count of 27578 vertices (noted as $27k$).

\begin{figure*}[t!]
\begin{center}
  \includegraphics[clip,width=\linewidth]{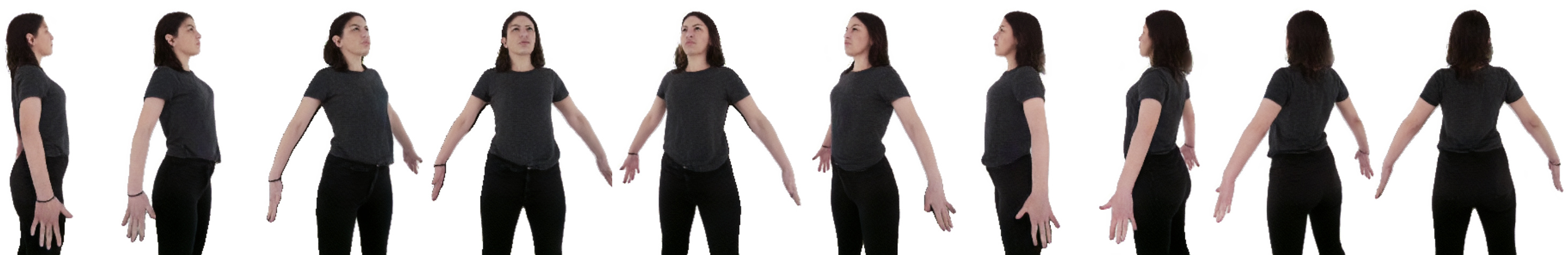}
\end{center}
 \caption{\textit{Novel view generation}. Images produced by our renderer are consistent across novel camera views.}
\label{fig:view}
\vspace{-5mm}
\end{figure*}

\textbf{Results.}
The values for PSNR and LPIPS are compiled in Table~\ref{table:results}. The first conclusion to extract from this table is that, given a fixed amount of color information (i.e.~comparing per-verts NMR against SMPLpix for a fixed topology), SMPLpix clearly outperforms NMR in both PSNR and LPIPs. Limiting the color information can be useful in terms of computational and data transmission efficiency, and the use of textures makes the rendering system arguably more complex.
However, we included also a comparison against NMR using full textures. Although the values are much closer, SMPLpix slightly outperforms NMR also in this case.
This validates our main hypothesis, i.e.~that the adaptive rendering procedure described in Section \ref{subsec:rasterization} can learn a valid rendering prior of the human texture and surface, and reproduce it based on a sparse input given by the colored mesh vertices. Moreover, it outperforms the conventional methods in terms of acquired level of realism since it is trained end-to-end to reproduce the corresponding photo. 
In terms of efficiency, using low-dimensional geometry with no anti-aliasing and full textures achieves the fastest running times (14ms), followed closely by SMPLpix (17ms), which obtains better quality metrics.  Also, note that for NMR, the inference time grows roughly linearly with the number of geometry elements, while for our method, most of the time is spent in the neural rendering module that is agnostic to the number of projected points. Being a UNet-like neural network, this module can be further optimised and tailored to specific hardware requirements.

\begin{figure*}[t!]
\begin{center}
  \includegraphics[clip,width=\linewidth]{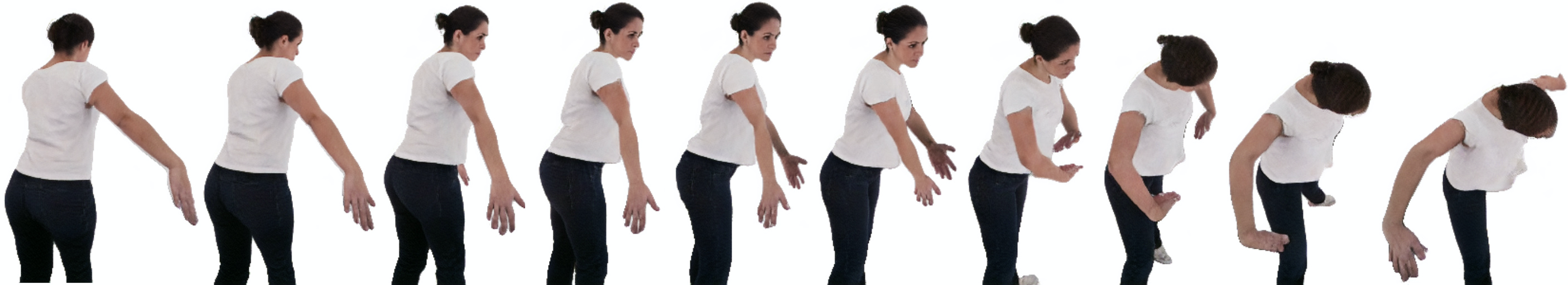}
\end{center}
 \caption{\textit{Pose generation}. We can animate subjects with novel pose sequences, e.g.~the ones taken from \cite{mahmood2019amass}. Please see the supplementary video on the project website for more examples of avatar reposing.}
\label{fig:motion}
\vspace{-5mm}
\end{figure*}

\begin{table}
\begin{center}
\caption{\emph{Neural mesh renderer \cite{kato2018neural} vs SMPLpix neural rendering pipeline.} Our model outperforms all variants of standard mesh rendering in both pixel-wise and perceptual similarity metrics.}
\label{table:results}
\begin{tabular}{l|c|c}
\hline\noalign{\smallskip}
 Method & PSNR $\uparrow$ & LPIPS $\downarrow$   \\
\hline
  NMR\cite{kato2018neural} (7k, per-verts) & 23.2 & 0.072 \\
  NMR\cite{kato2018neural} (7k, full textures) & 23.4 & 0.049 \\
  NMR\cite{kato2018neural} (27k, per-verts) & 23.5 & 0.064 \\
  NMR\cite{kato2018neural} (27k, full textures)& 23.6 & 0.047  \\
 SMPLpix (7k verts) &  24.2  & 0.051 \\
 SMPLpix (27k verts) & \textbf{24.6}  & \textbf{0.045} \\
 \hline
\end{tabular}
\end{center}
\end{table}

\subsection{Qualitative experiments}
\label{subsec:qualitative}

Since it is well known that perceptual metrics are not perfect in capturing the quality of synthetic images, we also provide examples for the reader to judge the quality of our method and to suggest the potential applications that its generative character enables. 

\textbf{Qualitative comparison.}
We provide a visual comparison of ground truth and the methods previously described in Figure~\ref{fig:fig2}. The first thing to note is that these images contain elements that are known to be difficult to model with the SMPL topology, e.g.~hair, baggy clothes, and shoes. We can observe that, since the relation between geometry and pixel colors in NMR is very constrained, the geometry artifacts are still visible in the rendered images. Note, for example, the unrealistic hair buns in NMR, smoothed out clothes in the first column, and the unrealistic ear shape in the sixth column due to the lack of independent hair geometry that covers the ears in the SMPL topology. In comparison, SMPLpix learns to correlate those artifacts with specific combinations of vertex locations and shapes, and recreates loose hair, pony tails, or loose clothing (to some extent). Another type of artifact that is corrected is incorrect texture due to misalignment: as seen in the fourth column, the hand area contain pixels of background color due to misalignment. SMPLpix learns to correct this type of artifact. Finally, pay attention to the toes rendered on the shoes by NMR, which are due to the SMPL topology. These toes  are corrected (removed) by our renderer in the next to last column. It is important to note that some of these details are reconstructed in a plausible way, though not in the exact way they are present in the ground truth.

\begin{figure}[b]
\begin{center}
  \includegraphics[clip,width=\linewidth]{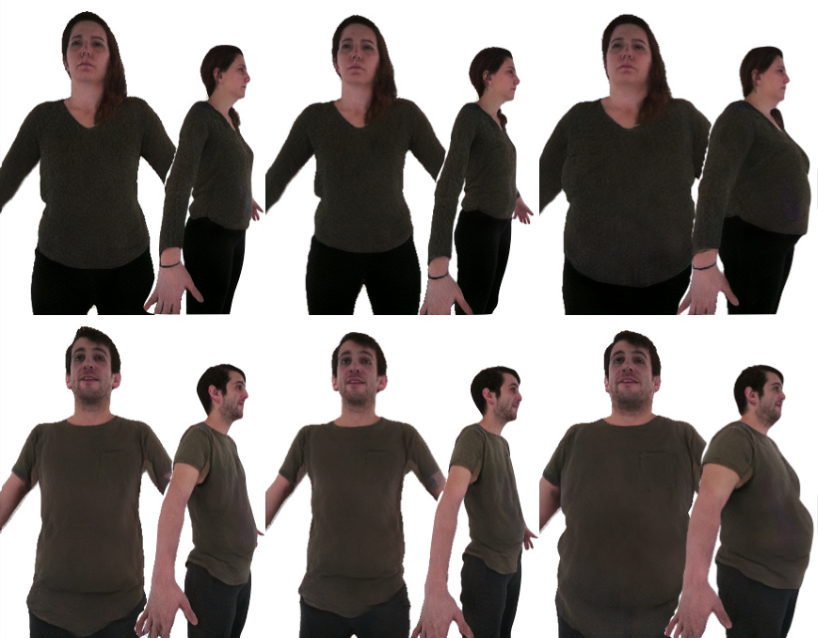}
\end{center}
 \caption{\textit{Shape variations with SMPLpix. The first column shows renderings of the original subject from two views. Subsequent columns explore the first directions of the SMPL shape space, in the negative and positive directions. This varies the subject shape, making them thinner or heavier, respectively.}}
\label{fig:shape}
\vspace{-5mm}
\end{figure}

\textbf{Novel view generation.}
A first question about SMPLpix generalization capabilities is how well does it generalize to novel views. Figure~\ref{fig:view} shows images generated from novel viewpoints with our algorithm. Given the ample coverage of views achieved by the scanning data, we can generate views from almost arbitrary orientations. However, we should note that the distance to the subject is not covered nearly as well in our setup, and the quality of our results degrade when the camera is too far or too close to the person. A possible way to handle this, left for future work, is to augment the data with arbitrary scalings of the input mesh and image.

\textbf{Pose generation.}
An advantage of our method with respect to the main other point-based renderer~\cite{aliev2019neural} is that we can alter the renders in a generative manner, thanks to the SMPL model that generates our inputs. To that end, we take the registrations previously mentioned and create a subject specific model in the same spirit as in~\cite{dyna}. A subject specific model has a template that is obtained by reverting the effects of the estimated registration pose. More specifically, it involves applying the inverse of the LBS transformation $W^{-1}$ and subtracting the pose-dependent deformations $B_P(\Vec{\theta})$ (Equations~\ref{eq:smpl_lbs} and ~\ref{eq:smpl_temp}) from the registration.
We can repose a subject specific model to any set of poses compatible with the SMPL model. To that end, we tried some sequences from AMASS~\cite{mahmood2019amass}. As can be seen in Figure~\ref{fig:motion}, bodies can deviate largely from the A-pose in which most of the subjects stand in the training data. Experimentally, we have observed that this is very different for other neural renderers like~\cite{chan2019everybody}.

\textbf{Shape generation.}
Although~\cite{aliev2019neural} cannot generate people arbitrarily posed, other renderers like~\cite{chan2019everybody,shysheya2019textured} potentially can, if they have a way to generate new skeleton images. However, shape cannot change with those approaches, since skeletons only describe the length of the bones and not the body structure. We can see this potential application in Figure~\ref{fig:shape}. For this figure, we used the previously mentioned subject-specific SMPL model for two of the subjects, and modified their shape according to the first three components of the original SMPL shape space. We can see that shape variations are realistic, and details like hair or clothing remain realistic. To our knowledge, this is the first realistic shape morphing obtained through neural rendering.

We provide more examples of avatar reposing, reshaping and novel view synthesis on the project web site\footnote{\url{https://sergeyprokudin.github.io/smplpix/}}.

\section{Conclusion and future work}

In this work, we presented SMPLpix, a deep learning model that combines deformable 3D models with neural rendering. This combination allows SMPLpix to generate novel bodies with clothing and with the advantages of neural rendering: visual quality and data-driven results. Unlike any other neural renderers of bodies, SMPLpix can vary the shape of the person and does not have to be retrained for each subject.

Additionally, one of the key characteristics of SMPLpix is that, unlike the classic renderers, it is improvable and extensible in a number of ways. We are particularly interested in integrating the renderer with systems that infer SMPL bodies from images (e.g.~\cite{kanazawa2018end,kocabas2019vibe,kolotouros2019learning}) to enable an end-to-end system for body image generation trained from images in the wild.

SMPLpix represents a step towards controllable body neural renderers, but it can obviously be improved. Rendering high-frequency textures remains a challenge, although including extra information in our input projection image is a promising approach; e.g.~per-vertex image descriptors, similar to the local image descriptors pooled across views in~\cite{pifu} or deep point descriptors in \cite{aliev2019neural}. 

\paragraph{Disclosure.}
While MJB is also an employee of the Max Planck Institute for Intelligent Systems (MPI-IS), this work was performed solely at Amazon where he is a part time employee. At MPI-IS he has received research gift funds from Intel, Nvidia, Adobe, Facebook, and Amazon. He has financial interests in Amazon and Meshcapade GmbH.

\clearpage

{\small
\bibliographystyle{ieee_fullname}
\bibliography{egbib}
}

\end{document}